# Single-Pixel Tactile Skin via Compressive Sampling

**Authors:** Ariel Slepyan[1]*†, Laura Xing[2]†, Rudy Zhang[1]†, Nitish Thakor[1,2,3]*

**Affiliations:**

[1]Department of Electrical and Computer Engineering, Johns Hopkins University, 3400 North Charles Street, Baltimore, MD 21218, USA.

[2]Department of Biomedical Engineering, Johns Hopkins School of Medicine, 720 Rutland Avenue, Baltimore, MD 21205, USA.

[3]Department of Neurology, Johns Hopkins University, 600 North Wolfe, Baltimore, MD 21205, USA.

*Corresponding author. Ariel Slepyan: aslepya1@jhu.edu. Nitish Thakor: nitish@jhu.edu

†These authors contributed equally to this work

**Abstract:** Development of large-area, high-speed electronic skins is a grand challenge for robotics, prosthetics, and human-machine interfaces, but is fundamentally limited by wiring complexity and data bottlenecks. Here, we introduce Single-Pixel Tactile Skin (SPTS), a paradigm that uses compressive sampling to reconstruct rich tactile information from an entire sensor array via a single output channel. This is achieved through a direct circuit-level implementation where each sensing element, equipped with a miniature microcontroller, contributes a dynamically weighted analog signal to a global sum, performing distributed compressed sensing in hardware. Our flexible, daisy-chainable design simplifies wiring to a few input lines and one output, and significantly reduces measurement requirements compared to raster scanning methods. We demonstrate the system's performance by achieving object classification at an effective 3500 FPS and by capturing transient dynamics, resolving an 8 ms projectile impact into 23 frames. A key feature is the support for adaptive reconstruction, where sensing fidelity scales with measurement time. This allows for rapid contact localization using as little as 7% of total data, followed by progressive refinement to a high-fidelity image – a capability critical for responsive robotic systems. This work offers an efficient pathway towards large-scale tactile intelligence for robotics and human-machine interfaces.

**One-Sentence Summary:** Compressed sensing enables scalable, high-speed tactile skins with a single output wire.

# Main Text:

## Introduction

Humans possess an exceptionally scalable and efficient tactile sensing system: approximately 230,000 mechanoreceptors, each about 10 μm in diameter, are distributed over roughly 2 m² of skin, with signal processing capabilities on the order of 1 ms precision [1], [2], [3], [4]. Replicating such a "human-scale" tactile sensing system remains one of the grand challenges in artificial tactile sensing.

Achieving this level of scalability has transformative potential across many domains. In robotics and prosthetics, it could enhance dexterity, adaptability, and environmental awareness in unstructured settings [5], [6]. In consumer electronics and human-computer interaction, it could lead to more immersive touchscreens and tactile interfaces [7], [8]. Moreover, it could enable emerging applications where tactile sensing is currently underutilized due to limitations in scalability and responsiveness.

Conventional tactile arrays, which typically use raster scanning for readout, are fundamentally limited in scalability. Raster scanning reads each sensor sequentially to prevent crosstalk [9], [10] resulting in significant latency as the number of sensors increases [11]. Furthermore, the wiring overhead – requiring one line per row and column – becomes prohibitive as array size and resolution scale up [12].

To address these challenges, recent research has explored "scalable" tactile arrays that reduce wiring complexity by sharing bus lines and transmitting data asynchronously based on detected events [13], [14], [15]. While these methods improve bandwidth efficiency, they rely on intelligent multiplexing (e.g., TDMA/CDMA) and often demand complex readout circuitry and high-speed processors to handle the data throughput.

An alternative approach is compressed sensing (CS), which offers the potential to reduce both hardware and computational requirements. CS works by linearly mixing sensor outputs into a small number of global measurements, exploiting signal sparsity to reconstruct the full signal with high fidelity. This paradigm has already shown success in applications like high-speed imaging [16] and low-power sensing systems [17]. However, applying CS to large-area tactile sensing has remained largely impractical. Traditional CS implementations require individual electrical access to each sensor element for encoding, which is incompatible with flexible, large-area tactile skins due to wiring and packaging complexity. As a result, most prior work in compressive tactile sensing remains limited to simulations [18], [19], [20], [21], [22] or very small-scale rigid setups [23], [24].

Despite these challenges, tactile signals are inherently well-suited for compressive sampling. Tactile interactions exhibit significant spatiotemporal sparsity and structured patterns [25], [26] which makes them highly compressible.

In this work, we present a scalable, single-output tactile sensing skin that enables distributed compressive sampling directly in hardware. By embedding analog weighting electronics into each sensing element using ultra-low-cost microcontrollers, our design enables each tactile pixel to contribute a weighted signal to a global output– implementing the compressive sensing matrix multiplication directly in the analog hardware with only a few input lines and a single output wire. The system is daisy-chainable, requires minimal wiring, and is compatible with flexible substrates.

We detail the sensor's architecture and signal reconstruction pipeline, and demonstrate its capability in high-speed object tracking, real-time contact localization, and rapid object classification. We also analyze the trade-offs between measurement rate, reconstruction accuracy, and speed, showing that high performance can be achieved with as little as 5–15% of the measurements required by conventional raster-scanning methods, enabling real-time tactile intelligence at scale.

## Results

*Single-Pixel Tactile Skin*

The Single-Pixel Tactile Skin (SPTS) utilizes compressed sensing to measure the tactile sensor array signal using a fraction of the measurements required to read each pixel individually, achieving a proportionally higher sampling rate (Figure 1).

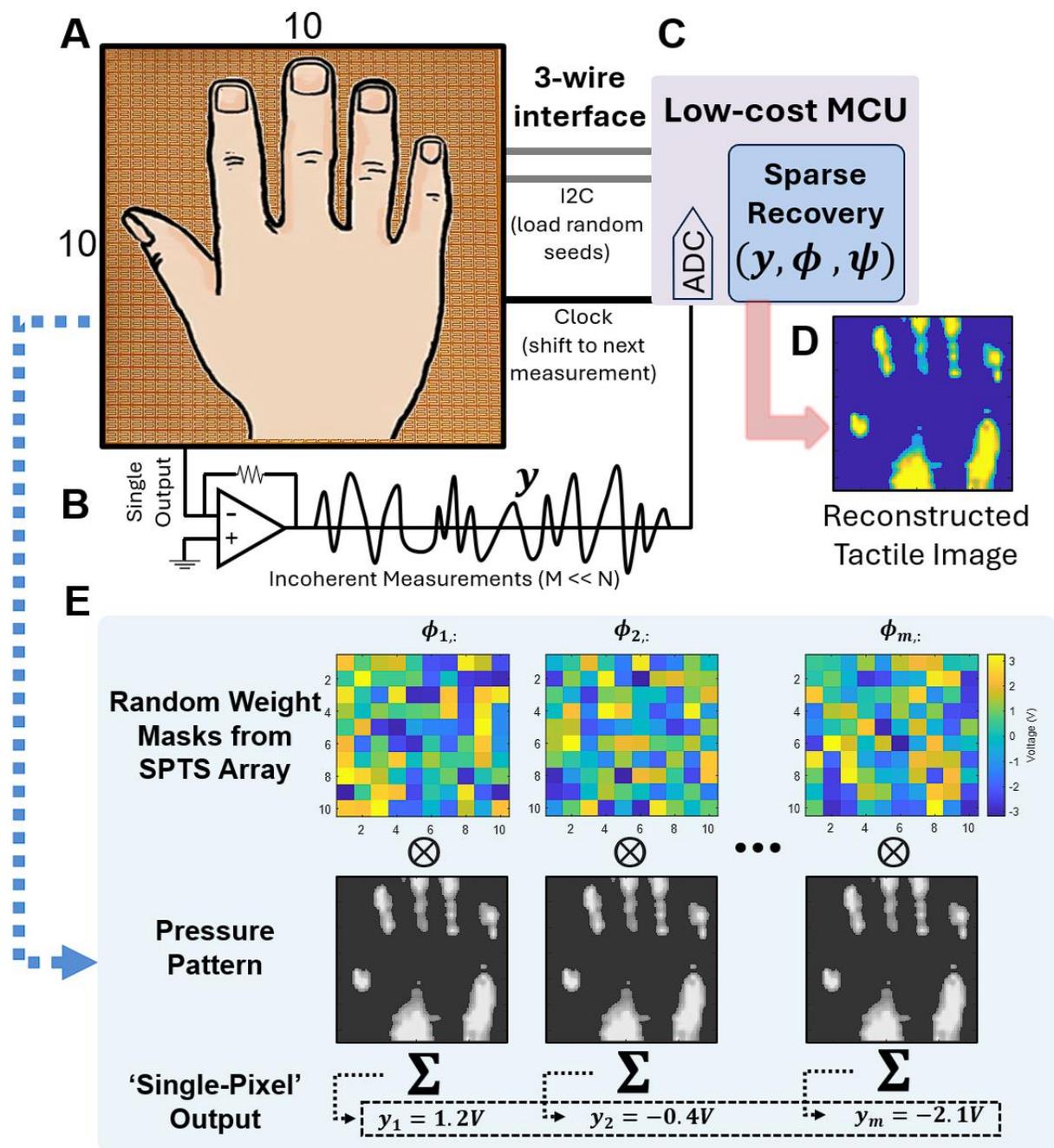

**Figure 1. Single-Pixel Tactile Skin. A.** Single-pixel tactile skin configured as a 10x10 tactile sensor matrix, with N=100 pixels. **B.** SPTS cells output a random projection of the tactile signal onto a single output wire. **C.** A low-cost, low-rate ADC captures the projection measurements over time obtaining a few measurements *M* that is significantly less than the total number of sensors N . **D.** Sparse recovery algorithm calculates the reconstructed tactile image with N pixels using the projection matrix used for obtaining y measurements and a pre-collected tactile dictionary. **E.** The conceptual process of SPTS, where the tactile pressure pattern is multiplied by a series of pseudo-random weight masks (Φ). Each multiplication yields a 'single-pixel' summed output (y), representing a unique random projection of the image. A small number of these successive measurements (M<<N) are then used to reconstruct the entire tactile image.

Compressed sensing is achieved by dedicating a low-cost small footprint microcontroller for each pixel of the tactile array and programming it to output a sequence of random analog voltages defined by a pre-loaded random seed which is loaded by a globally shared I2C bus. A shared clock signal, which sets the overall measurement rate of the system, tells the sensors when to progress to their next analog voltage value.

The random analog voltage works as a driving 'weight' for that sensing pixel, allowing us to sum all the tactile signals with according to a typical compressed sensing problem. To sum the signals according to the compressed sensing principle an inverting amplifier circuit is used setup as a voltage summer. This allows us to formulate the problem as solving for the resistance of each pixel using circuit analysis as follows:

$$V_{Out} = -R_f \left( \frac{V_1}{R_{S1}} + \frac{V_2}{R_N} + \cdots + \frac{V_N}{R_{SN}} \right)$$

where $R_f$ is the resistance of the feedback resistor in the summing amplifier, $V_i$ is the analog weighting voltage for each sensor, and $R_{Si}$ is the resistance signal we attempt to measure. As evident from the equation, each measurement of $V_{Out}$ contains information about all the resistance values in the entire tactile skin. By randomly varying the driving voltages $V_i$, and collecting several measurements of $V_{Out}$, several diverse projections of the $R_{Si}$ vector are obtained. This yields the sensor data acquisition formulation of $M$ measurements of $V_{Out}$ as:

$$\begin{bmatrix} V_{O1} \\ V_{O2} \\ \vdots \\ V_{OM} \end{bmatrix} = \begin{bmatrix} V_{1,1} & \cdots & V_{N,1} \\ \vdots & \ddots & \vdots \\ V_{1,M} & \cdots & V_{N,M} \end{bmatrix} \begin{bmatrix} C_{S1} \\ C_{S1} \\ \vdots \\ C_{SN} \end{bmatrix} = y = \phi x$$

where the conductivity is replaced for the inverse of resistance as $C_{Si} = 1/R_{Si}$. The voltage matrix operates as the projection matrix $\phi$ and the conductance vector represents the signal we desire to reconstruct $x$. Following this formulation, the reconstructed signal $\hat{x}$ can be reconstructed using a sparse recovery algorithm and a pre-acquired tactile dictionary as $\hat{x} = SR(\phi\psi, y)$. Circuit implementation is shown in Figure 2B.

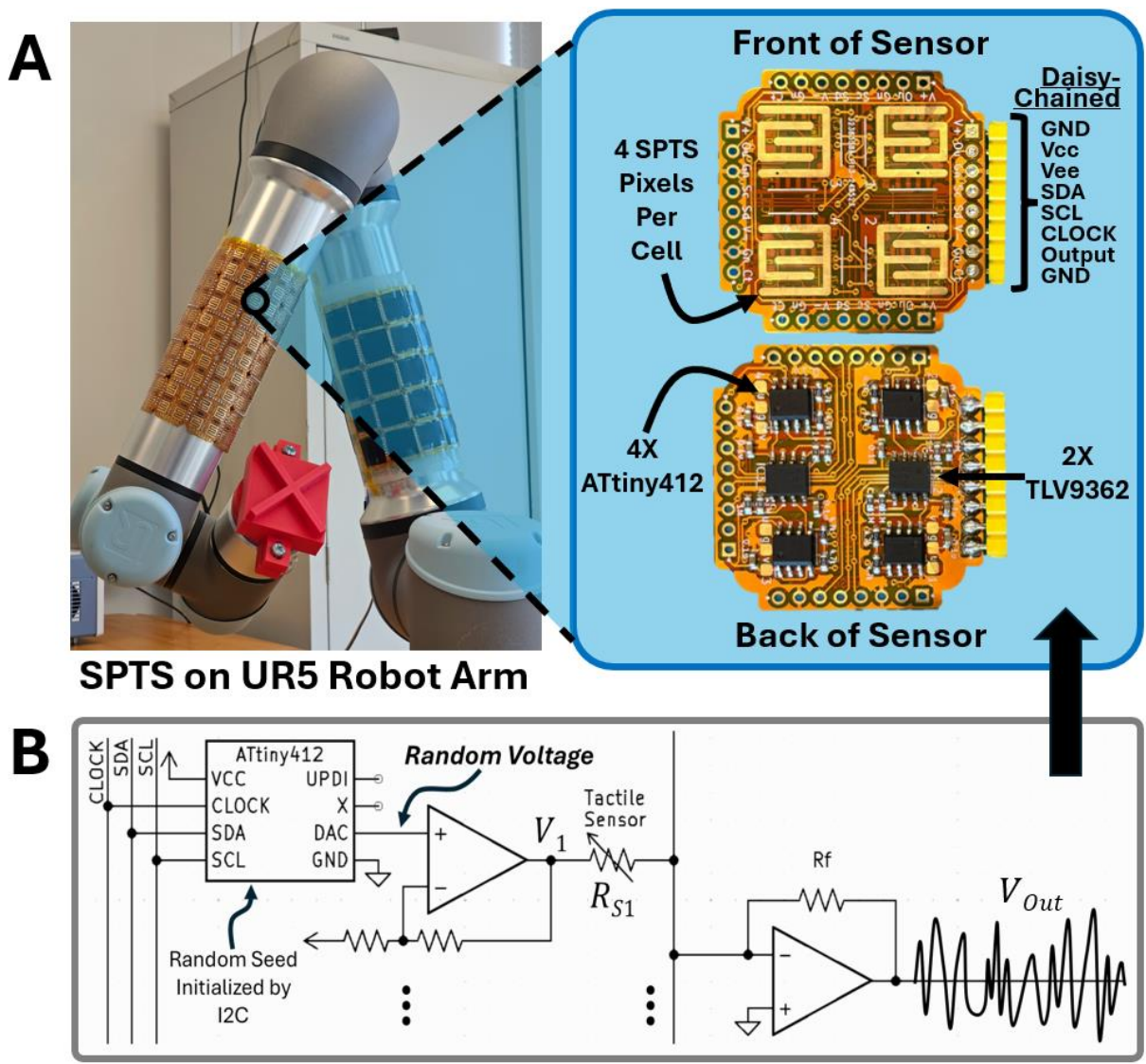

**Figure 2. SPTS Sensors and Readout. A.** SPTS covering the UR5 robotic arm. Zoom-in shows the front and back of a single 'cell'. **B.** Each pixel of SPTS uses a low-cost, small-footprint microcontroller (ATtiny412) to produce a pseudorandom driving voltage for each sensing pixel. An amplifier circuit centers the voltage at 0V to ensure the voltage sum does not contain a high DC offset. The output of each tactile sensor is wired together, and sensor output is combined using a summing circuit. SPTS cells are daisy-chained together to share an I2C bus for loading random seeds and share a 'CLOCK' signal that cycles the cells to proceed to the next random driving voltage signal.

## Rapid Tactile Reconstruction and Classification of Daily Objects

The reduced number of measurements ($M$) inherent to the SPTS design directly translates to accelerated object reconstruction and classification without a significant compromise in accuracy. To quantitatively evaluate these capabilities, we assessed SPTS performance using a diverse library of 17 objects, encompassing common household items and 3D-printed geometric shapes. As depicted in Figure 3A, these objects were systematically indented into the SPTS sensor array by a robotic arm (Universal Robots UR5 Robot). For each object, 10 indentation trials were conducted, acquiring data in SPTS mode across various compressive measurement levels ($M$) and, for control, using a traditional raster-scan mode.

First, we evaluated the fidelity of tactile image reconstruction. Full tactile images were reconstructed from the SPTS compressive measurements using a pre-learned tactile dictionary (see Methods and Supplement S6 for dictionary details). To quantify geometric reconstruction quality, we calculated the "support accuracy." This metric assesses the accuracy of identifying the pixels in contact with the object, thereby reflecting the correctness of the reconstructed shape. Figure 3C presents the average support accuracy for SPTS across all 17 objects as a function of $M$. For fair comparison, these results are benchmarked against equivalent support accuracy derived from raster-scanned images that were down-sampled to equivalent $M$ levels and interpolated to match the effective data output corresponding to each $M$ level of SPTS. For smaller objects (contact area < 40% of total pixels), SPTS $M$ values as low as 15-20 yielded sufficiently accurate reconstructions. Larger objects, as expected, benefited from higher $M$ values, typically in the range of 25-30, to achieve comparable support accuracy.

Next, we assessed the system's ability to classify objects based on the reconstructed tactile images. Each SPTS reconstruction was fed into a Sparse Representation-based Classifier (SRC) trained on a library of tactile images of the 17 test objects (classifier details in Methods). Figure 3B shows the average classification accuracy for SPTS across all objects as $M$ varies. Similar to the reconstruction evaluation, the raster-scanned signals used for comparative performance points in Figure 3B were appropriately down-sampled and interpolated. The SPTS system achieved ≥98% classification accuracy using just $M=20$ measurements, corresponding to an effective frame rate of 3500 FPS. Even with as few as $M=13$ measurements (5400 FPS), an average accuracy of at least 80% was maintained.

Finally, to highlight the practical speed advantage of SPTS for time-critical applications, we conducted a rapid classification test, measuring the time from the start of data acquisition to successful object identification upon initial contact. As shown in Figure 3D, SPTS operating with $M=25$ measurements can classify an indented object with 80% accuracy within just 0.4 ms of data collection. In contrast, a traditional raster scan (equivalent to $M=N=100$ where $N$ is the total pixel count) required at least 1.6 ms – 4 times longer – to acquire the first complete tactile frame necessary for comparable classification, underscoring the significant latency reduction offered by our compressive approach.

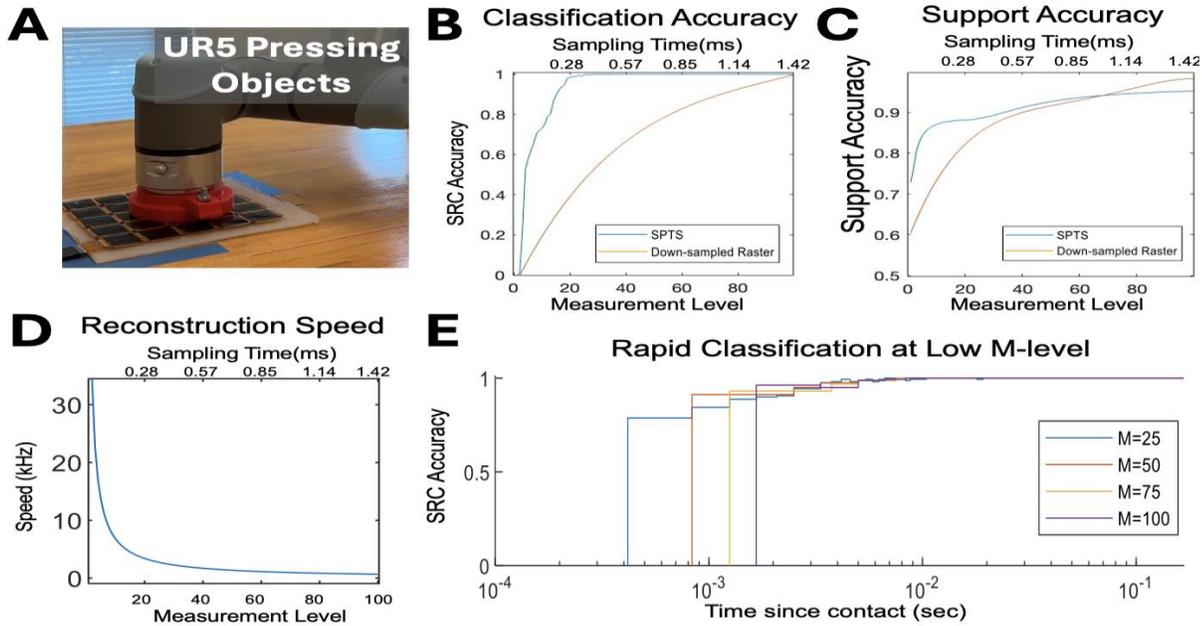

**Figure 3. Rapid Tactile Reconstruction and Classification. A.** UR5 indented 17 different daily and 3D printed objects in the tactile sensor array. Data was collected using SPTS-mode at different measurement levels and also raster-scan mode as a control. **B.** Overall accuracy of SPTS in object classification. Raster-scanned signal was down-sampled and interpolated to achieve equivalent comparison points. **C.** Overall accuracy of SPTS in reconstructing the support of the indented object compared to equivalent interpolated down-sampled raster-scan image. **D.** Frequency of reconstruction, in kilohertz, using different measurement levels for SPTS. **E.** Rapid classification of object at initial contact using low measurement levels. $M=25$ can classify the indented object within 0.4msec of data collection with 80% accuracy.

## *Adaptive Reconstruction for Progressive Refinement*

A key advantage of the SPTS framework is its inherent support for adaptive reconstruction. By sequentially acquiring and incorporating additional compressive voltage measurements into the reconstruction process, the fidelity of the tactile image can be progressively enhanced. This allows for a hierarchical approach to tactile perception: initial, coarse features such as contact location and overall shape can be rapidly estimated using a small number of measurements, while finer details can be resolved by subsequently processing more measurements. This dynamic trade-off between speed and detail is advantageous for real-world robotic applications requiring both rapid reflexes and detailed object analysis.

To qualitatively demonstrate this adaptive capability, various objects were indented onto the SPTS array using a robotic arm (Universal Robots UR5 Robot). The tactile images were reconstructed iteratively, incorporating an increasing number of compressive measurements $M$. Figure 4 specifically illustrates this iterative refinement for a "T" shaped object (shown as an inset on the sensor in Figure 4) indented onto the SPTS. The figure displays a sequence of 15 distinct voltage measurements and shows the corresponding tactile image reconstructions calculated using progressively more of these measurements–starting with reconstructions based on only the first 2 measurements and culminating in a reconstruction using all 15 available measurements. As indicated by the colored groupings in Figure 4, which associate specific sets of measurements with their reconstructions, the visual quality and detail of the reconstructed "T" shape improve as more measurements are included in the computation. For example, while a reconstruction with very few measurements (e.g., $M=2$ or $M=5$) might only reveal the most prominent contact points or a blurred outline, by the time $M=15$ measurements are utilized, the "T" shape becomes distinctly recognizable, with clearer edges and more accurate intensity distributions. The exact number of measurements for convergence to a high-fidelity image varied with object size and complexity.

The dynamic nature of adaptive reconstruction is further showcased in Video V1. In this demonstration, a tennis ball is dropped onto the SPTS. The resulting tactile event is reconstructed iteratively: first using only the initial measurement, then the first two, and so on, up to a total of $M=N=100$ measurements. The video depicts the reconstructed frame evolving from subtle, noisy signals to revealing distinct high-intensity pixel estimates in the central contact area, ultimately forming a recognizable tennis ball shape by approximately $M=15$ measurements. This highlights the system's ability to track dynamic events with progressively improving spatial resolution as more data becomes available within a very short timeframe.

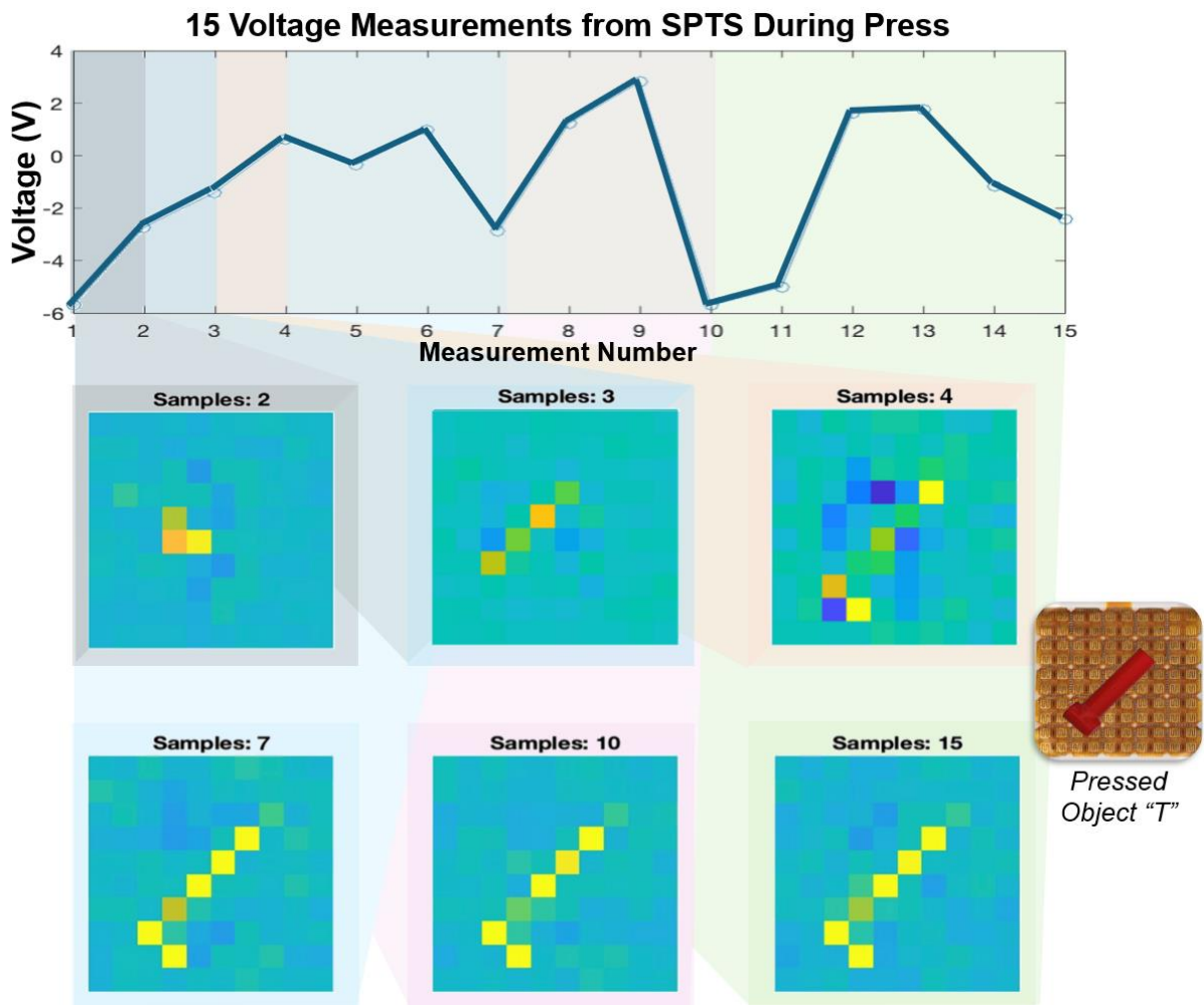

**Figure 4. Adaptive Reconstruction.** As additional voltage measurements are recorded, the reconstruction accuracy of the tactile image improves. 15 voltage measurements are shown, and tactile image reconstructions are calculated using progressively more of the measurements: with only the first 2 measurements, through using all 15 measurements. Measurements and their reconstructions are associated with colored groupings. The indented object, the shape "T", is shown as an inset on the sensor.

## High-Speed Projectile Tracking

The significantly enhanced effective frame rate achieved by SPTS enables the detailed capture of transient, high-speed tactile events, such as object collisions. To demonstrate this capability, we performed a high-speed impact test, tracking the dynamics of a tennis ball dropped onto the sensor array (Figure 5A). This experiment serves as a model for characterizing brief, transient contact events, which are relevant to applications such as collision detection, object catching, and dexterous manipulation in robotics. The entire contact duration for a single bounce was very brief, lasting approximately 8 ms.

Using our compressive reconstruction scheme with $M=25$ measurements per frame, SPTS successfully captured an average of 23 distinct tactile frames throughout this 8 ms bounce event. This high temporal resolution is critical for accurately characterizing impact dynamics. Conversely, acquiring more measurements per frame inherently takes longer. This increased acquisition time per frame results in fewer total frames captured during the brief contact event, leading to lower temporal resolution and, consequently, larger average pressure changes (Δ Pressure) observed between consecutive frames (Figure 5B).

The impact of this temporal resolution on tracking rapid events is clearly shown in Figure 5C, which plots the evolution of maximum pressure during a tennis ball bounce for different measurement levels. SPTS operating at lower measurement levels (e.g., $M=25$) detects the initial impact and subsequent pressure changes with finer granularity and often earlier in the event timeline compared to higher measurement levels like $M=100$ (full raster scan). For instance, Figure 5B illustrates that the smoother, more gradual changes in pressure during the collision are better captured with the higher frame rates afforded by lower $M$ values. This finer temporal detail can be crucial for understanding precise mechanics of a dynamic interaction.

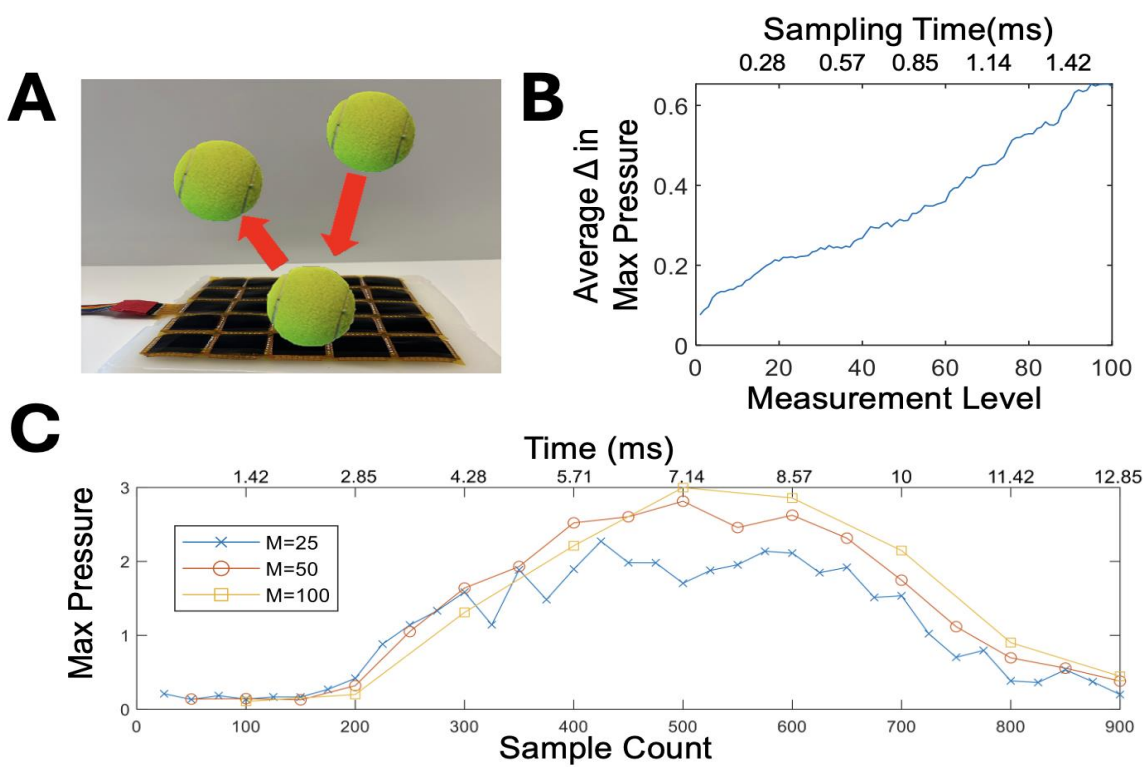

**Figure 5. High-Speed Projectile Tracking. A.** Tennis ball is bounced on the sensor array, making contact with the array for only ~8msec. **B.** Lower measurement levels capture the dynamic signal with higher accuracy due to higher frame rates, and the average change (Δ) in pressure value is plotted. Higher measurement levels have lower frame rates and accordingly have larger average pressure changes per frame. **C.** Evolution of maximum pressure during a tennis ball bounce versus measurement level. Lower measurement levels ($M=25$) 'notice' the bounce of the ball earlier, and track the deformation pressure with finer granularity than higher measurement levels ($M=100$).

*Rapid Contact Localization*

Beyond capturing high-speed dynamics, the tennis ball impact experiments also highlight SPTS's capability for rapid contact localization. Accurate and fast localization is often a prerequisite for subsequent, more detailed object analysis or for initiating reactive behaviors in robotic systems. Using the SPTS reconstruction scheme, we calculated the center of mass (CoM) of the reconstructed tactile image from the tennis ball impacts and compared this estimated CoM to the ground truth center of contact. This comparison was performed across varying numbers of compressive measurements ($M$) to assess the trade-off between measurement count and localization accuracy (Figure 6).

The results demonstrate that a robust estimate of contact location can be achieved with very few measurements, fewer than those required for high-fidelity full object shape reconstruction. As shown in Figure 6, the average distance between the estimated CoM and the true contact location decreases as $M$ increases. Notably, even with as few as 7 measurements, the object's location can be estimated with an error consistently within a 2-pixel radius of the true center. By approximately $M=10$ measurements, our localization estimate is typically adjacent to the actual object location. The localization accuracy continues to improve with additional measurements, reaching a point where the estimated contact center is, on average, within 1 pixel of the actual contact center by $M=25$ measurements. Beyond this point, further increases in $M$ yield diminishing returns in localization accuracy, as indicated by the plateau in Figure 6.

This ability to achieve rapid and reasonably accurate contact localization with a minimal number of measurements (e.g., $M$ in the range of 7-15) underscores the efficiency of SPTS for applications demanding quick initial assessments of tactile interactions, such as determining where contact has occurred before committing resources to more detailed reconstruction or classification.

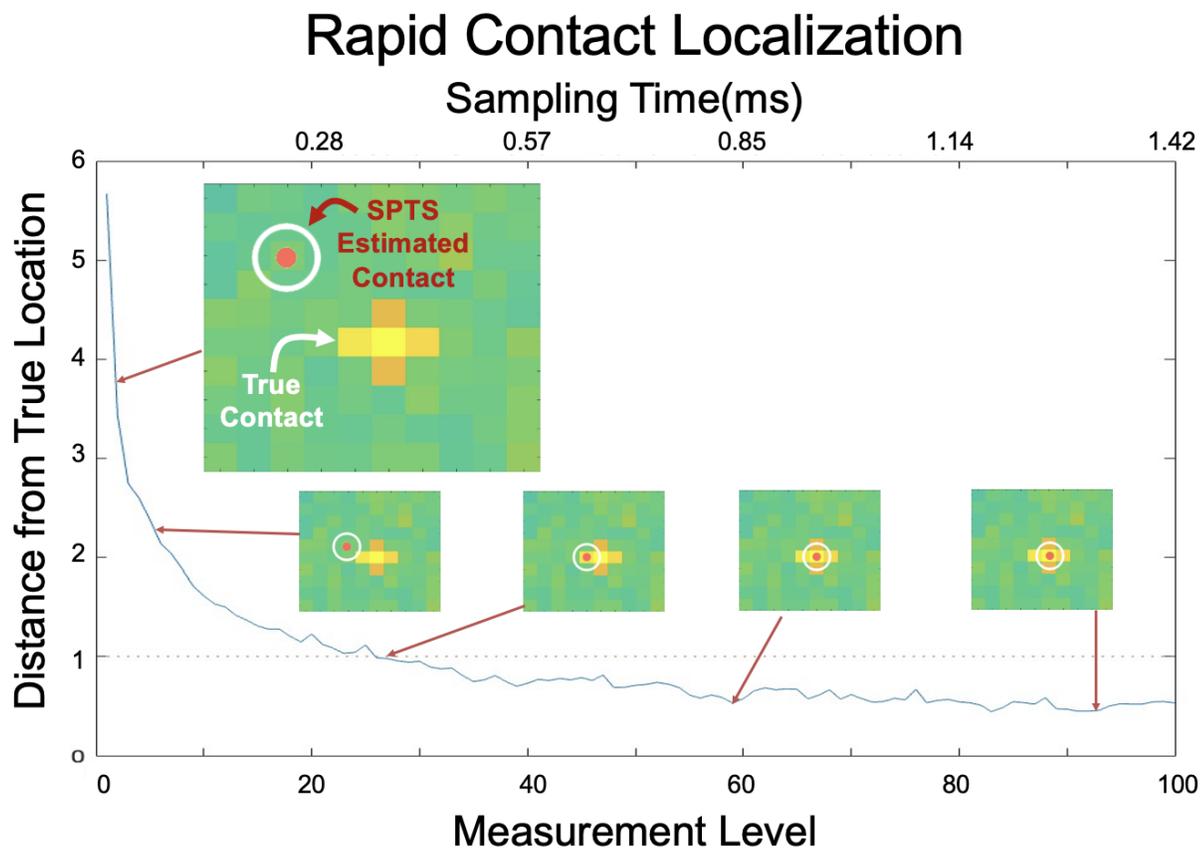

**Figure 6. Rapid Contact Localization.** Center of contact can be estimated with fewer measurements than accurate object reconstruction requires. During various tennis ball bounces, the average distance from true contact location was estimated using a low measurement level. As the level increase, the contact localization accuracy improves and plateaus after 25 measurements. Within 7 measurements, the objects location can be accurately estimated within span of 2 pixels error.

## DISCUSSION

This work demonstrates a significant advance in tactile sensing by realizing a scalable, high-speed system based on in-sensor analog compressive sensing (CS). A key innovation lies in the direct circuit-level implementation of the CS framework for a resistive sensor array, achieved by employing distributed analog voltage weighting at each pixel and current summation via an operational amplifier, mathematically framed as a linear system solvable for pixel conductances. Our Single-Pixel Tactile Skin (SPTS) architecture, embodying this distributed principle, overcomes prior limitations in applying CS to practical, large-area tactile arrays. SPTS achieves accurate tactile image reconstruction, object classification, and dynamic event tracking using substantially fewer measurements (5-15%) than conventional methods, thereby enabling proportionally higher effective operational speeds and reduced data latency.

The successful hardware realization of CS for resistive arrays addresses a critical challenge in tactile technology: the trade-off between sensor coverage/density and system complexity, particularly wiring overhead and data bandwidth [11], [12]. Unlike existing approaches that rely on complex multiplexing or event-driven schemes for resistive arrays [13], [14], [15], or CS implementations limited to simulation or non-resistive modalities [18], [19], [20], [21], [22], [23], [24], SPTS offers an intrinsically streamlined data acquisition pathway rooted in fundamental circuit theory. This advance is significant for fields like robotics and prosthetics, where the development of large-area, human-like tactile skins has been a long-standing goal.

Crucially, the SPTS framework's reliance on a drastically reduced number of measurements distinguishes it from other sensing paradigms such as certain neuromorphic or event-based systems [13], [14], [15]. While those systems excel at capturing rich temporal details, they typically necessitate very high-speed readout circuitry and significant bandwidth for each sensor or event to resolve their encoded signatures uniquely. This demand for high-bandwidth data acquisition per element can ultimately limit scalability and impose significant processing burdens, especially for large arrays. SPTS, by contrast, performs inherent data compression at the physical layer via structured dimensionality reduction, synchronously acquiring global measurements and inherently reducing these downstream requirements. This reduction in hardware complexity and data load is useful for developing extensive tactile interfaces, such as those required for advanced robotics or comprehensive prosthetic feedback [5], [6].

The performance metrics achieved by SPTS underscore its potential. High-fidelity object classification (≥98% accuracy with $M=20$ measurements, ~3500 FPS) and sub-millisecond classification latency (80% accuracy within 0.4 ms) are critical for real-time robotic interaction and control, enabling potential applications such as rapid object identification or adaptive grip adjustment in robotic systems. The system's capacity to track transient events with high temporal resolution, exemplified by capturing ~23 frames of an 8 ms projectile impact, provides a level of dynamic detail previously inaccessible with comparable array sizes using raster scanning. Such data, derived from the efficient compressive sampling of pixel conductances, is essential for understanding complex contact mechanics and enabling rapid physical interactions. This capability demonstrates the system's suitability for future work in monitoring dynamic contact events, such as detecting incipient slip or characterizing tool-surface interactions in robotic manipulation.

Furthermore, the adaptive reconstruction capability, inherent to the SPTS framework, allows for dynamic adjustment of sensing fidelity, offering a powerful advantage for advanced robotics. This is particularly advantageous for tasks like rapid contact localization. In conventional array scanning, detecting contact, especially if sparse (e.g., activating only a single pixel), necessitates interrogating potentially all $N$ pixels, imposing a latency proportional to $N$. In contrast, SPTS achieves rapid contact localization (e.g., <2-pixel error with $M=7$ for an $N=100$ array) with significantly fewer measurements ($M \ll N$). This efficiency stems directly from the CS principle: each global measurement $V_{Out,i}$ contains weighted information from all pixels simultaneously. Therefore, even a small number of compressive measurements provides sufficient information to infer the approximate location of contact across the entire array, long before a full scan could be completed. This hierarchical information processing allows for a paradigm where a robot could, for example, prioritize initial contact localization to secure a grip, followed by subsequent refinement to a higher-resolution tactile image by acquiring and processing additional measurements (M=25 for <1-pixel localization error) to discern finer object features. This hierarchical information processing mirrors efficiencies found in biological sensory systems [27], [28] and offers a flexible paradigm for resource-constrained robotic perception. The potential to refine reconstructions by cumulating measurements, as suggested by the improved classification with $M$=50 at t=0.8ms, points towards implementations of "anytime" sensing algorithms, where information quality improves continuously with available sensing time or measurement count, a valuable feature for dynamic decision-making.

Future advancements can build upon this foundation. While SPTS provides significant advantages in speed and scalability, a current limitation related to the prototype's pixel density is the somewhat "blocky" nature of reconstructed shapes, reflecting the underlying resolution of the current sensor array prototype (1.5cm between taxels). Addressing this through further monolithic integration via ASICs or thin-film technologies could yield highly conformable arrays with significantly increased pixel densities and thus improved spatial resolution of reconstructions, it is important to note that even the current resolution demonstrates effectiveness for certain large-area tasks where precise edge definition is less critical than, for example, distributed contact localization or gross shape identification. Optimization of the CS process itself, through learned sensing matrices [29], [30] or advanced reconstruction algorithms including deep learning methods [31, 32], may further enhance efficiency and robustness. Extending the SPTS CS principle to other tactile transduction modalities that can be similarly mapped to a weighted summation (e.g., certain capacitive, piezoelectric configurations) and rigorously characterizing noise propagation in the shared analog pathway for very large-scale implementations are also important research directions.

In summary, SPTS provides a novel and effective solution for high-performance, scalable tactile sensing through the direct hardware implementation of compressive sensing, specifically tailored at the circuit level for resistive sensor arrays. By substantially mitigating wiring and data bottlenecks via its inherent compressive sensing encoding strategy, this approach offers a clear path towards creating sophisticated artificial tactile systems with capabilities essential for next-generation robotics, prosthetics, and human-machine interaction, moving closer to the efficiency and richness of biological touch.

## MATERIALS AND METHODS

### Sensor Construction

The Single-Pixel Tactile Skin (SPTS) was constructed using modular, flexible printed circuit board (FPCB) units. Each FPCB unit hosts four individual sensor pixels. A single sensor pixel comprises an ATtiny412 microcontroller (Microchip Technology) and a TLV9362 operational amplifier (Texas Instruments) with associated passive components. This op-amp circuit is configured to amplify, level-shift, and directly drive the sensor element with the microcontroller's processed DAC output. The microcontroller is programmed to generate unipolar analog voltages (0 to 3.3V) via its DAC. This DAC output serves as the input to the op-amp circuit, which transforms it into a bipolar weighting voltage spanning approximately -3.3V to +3.3V. This transformation is crucial to ensure that the sequence of pseudorandom weighting voltages has a mean close to zero, preventing DC offset accumulation in the subsequent analog summation stage, and to utilize a wider dynamic range for encoding.

The sensing element itself is integrated on the reverse side of the FPCB with interdigitated copper electrodes. These electrodes are then uniformly coated with a piezoresistive layer of Velostat (3M).

Mechanical pressure applied to this Velostat layer alters its local resistance over the interdigitated electrodes, forming the basis of tactile transduction.

Multiple FPCB units are interconnected to form the 10x10 pixel array (*N*=100) used in this study. Interconnections are facilitated by shared bus lines for power (VCC, GND), I2C communication (SDA, SCL) for initial seed distribution and control, and a global clock signal (CLOCK) to ensure synchronous operation of all microcontrollers during the generation of weighting voltages. This modular, daisy-chainable design allows for scalable sensor arrays with robust electrical connections and mechanical flexibility. The overall system architecture is detailed in Figure 1 and Figure 2.

*Pseudorandom Weighting Voltage Generation*

Each of the N=100 sensor pixels in the array is assigned a unique initial seed value for its pseudorandom number generator (PRNG), distributed via the I2C bus during system initialization. To generate the sequence of bipolar analog weighting voltages (designed to have a near-zero mean over the sequence and span a range of approximately +/- 3.3V), each microcontroller generates a unipolar pseudorandom value using a linear congruential generator (LCG) defined by:

$$Seed_{n+1} = a \times Seed_n + c$$

$$a = 1664525$$

$$c = 1013904223$$

$$unipolar\ voltage = seed/16777216$$

This Unipolar Output (ranging 0 to 3.3V) is then converted by the DAC and serves as an input to the per-pixel op-amp circuit. This circuit scales and shifts the unipolar DAC output to produce the final bipolar weighting voltage (approximately -3.3V to +3.3V) for pixel k at measurement m. This ensures that the ensemble of generated weights has a mean approaching zero. The global clock signal ensures all N microcontrollers advance their LCGs and their respective op-amp circuits output these values simultaneously to drive the sensor elements. This results in an *M* x *N* sensing matrix Φ whose elements are drawn from a distribution with a mean near zero over this bipolar range.

*Dictionary Learning*

Using the data collected from pressing various objects into the sensor array, we began by preprocessing the dataset to ensure that only meaningful, non-redundant signals were used for dictionary learning. This involved filtering out low-amplitude signals by applying a voltage threshold, which removed noise and unintentional contacts. We also implemented a coherence-based filtering step to eliminate data samples that were too similar to one another, ensuring a diverse and representative set of inputs. After isolating the unique and informative data, we applied the K-means Singular Value Decomposition (K-SVD) [31] algorithm to learn a dictionary optimized for sparse representation. The algorithm iteratively alternates between sparse coding and dictionary update steps to extract a set of basis vectors that can efficiently represent our sensor signals. We specified a target sparsity level of 30 and dictionary size of 100 through empirical testing.

*Sparse Recovery and Tactile Reconstruction*

Given *M* compressive measurements *y* and the sensing matrix Φ, the tactile signal *x* (representing scaled conductances, where $x_i = -R_f * C_{Si}$, was reconstructed by finding its sparse representation α in the learned dictionary Ψ ($x = \Psi\alpha$). Orthogonal Matching Pursuit (OMP) was used, terminating when a target sparsity *S* = *M*/4 (rounded to the nearest integer) was reached. The reconstructed tactile signal is then $\hat{x} = \Psi\hat{\alpha}$. Individual pixel conductances $C_{Si}$, are obtained by $C_{Si} = -\hat{x}_i/R_f$.

*Robotic Indentation with Object Library*

To systematically characterize the sensor performance, we developed a robotic indentation setup paired with a library of 17 objects (Supplemental Figure S5). The robotic arm applies controlled presses on the sensor array using objects of varying shapes, materials, and sizes from the library. This approach enables the generation of a comprehensive and labeled dataset that captures the distinct voltage responses of the sensor array to different contact profiles. The resulting dataset was used for dictionary learning, classification, and sparse reconstruction tasks, enabling the sensor system to recognize and differentiate objects based on tactile features.

*Object Classification*

To classify objects based on pressure data, we use a Sparse Representation-based Classification (SRC) framework combined with voting logic. For each press event, corresponding to a unique object, we extract a defined frame range of sensor measurements and their associated sensing matrices from our pseudorandom table. Each measurement segment is processed by first performing sparse recovery using Orthogonal Matching Pursuit (OMP), where we solve for sparse coefficients that best reconstruct the signal from the learned dictionary. The reconstructed signal is then classified using a nearest-classifier algorithm, which compares it to a pre-labeled object library. Classification is repeated across multiple overlapping reconstruction windows, and a majority-voting mechanism is applied every fixed number of reconstructions (e.g., every 20 frames) to determine the most likely object label. This classification process is iterated across a range of sampling levels and presses, and accuracy is averaged over all repetitions to assess system performance. The result is a robust, frame-wise object recognition algorithm that leverages both sparsity and temporal voting to improve classification reliability from random tactile input.

*Support Accuracy Evaluation*

Support accuracy was calculated as the percentage of correctly identified contact pixels. A ground-truth contact map (from raster scan) and a reconstructed contact map (from SPTS reconstruction) were binarized using an empirically determined threshold chosen to visually segment clear contact regions from background noise. These binary maps were then compared pixel-wise. This was averaged across objects and trials for Figure 3C. Comparative raster scan data in Figure 3C was processed using the same sparse pixel selection method described for object classification to match the $M$ value of SPTS, followed by interpolation back to the original array size before thresholding and comparison.

*High-Speed Object Tracking*

To evaluate the temporal resolution of our tactile sensing system, we performed high-speed object tracking using a tennis ball bouncing on the sensor array. We accurately localize the ball's point of contact using only minimal amount of data at high speed–approximately five consecutive frames. Despite the limited temporal data, the system reliably identifies the activated sensor region corresponding to the impact point, enabling precise spatial reconstruction of the ball's bounce. This experiment demonstrates the system's capability for fast, frame-efficient tracking of dynamic contact events, highlighting its potential for real-time tactile monitoring in applications like robotic interaction or smart surfaces.

**Supplementary Materials**

S1.  SPTS array assembly

S2.  SPTS cell schematic and board layout

S3.  Component list

S4.  Example tactile reconstructions of SPTS

S5.  Picture of the 17 objects

S6. Learned Tactile Dictionary

V1. Video of tennis ball impacting sensor and spatiotemporal tactile reconstruction

V2. Video of interacting with SPTS mounted on UR5

Videos can be accessed at this link: SPTS_videos

Code and data can be accessed at our GitHub

# References

[1] G. Corniani and H. P. Saal, "Tactile innervation densities across the whole body," *Journal of Neurophysiology*, vol. 124, no. 4, pp. 1229–1240, 2020, doi: 10.1152/jn.00313.2020.

[2] I. Moll, M. Roessler, J. M. Brandner, A.-C. Eispert, P. Houdek, and R. Moll, "Human Merkel cells – aspects of cell biology, distribution and functions," *European Journal of Cell Biology*, vol. 84, no. 2–3, pp. 259–271, Mar. 2005, doi: 10.1016/j.ejcb.2004.12.023.

[3] W. Montagna, A. M. Kligman, and K. S. Carlisle, "Skin Sensory Mechanisms," in *Atlas of Normal Human Skin*, New York, NY: Springer New York, 1992, pp. 191–223. doi: 10.1007/978-1-4613-9202-6_6.

[4] E. L. Mackevicius, M. D. Best, H. P. Saal, and S. J. Bensmaia, "Millisecond Precision Spike Timing Shapes Tactile Perception," *J. Neurosci.*, vol. 32, no. 44, pp. 15309–15317, Oct. 2012, doi: 10.1523/JNEUROSCI.2161-12.2012.

[5] H. Yousef, M. Boukallel, and K. Althoefer, "Tactile sensing for dexterous in-hand manipulation in robotics—A review," *Sensors and Actuators A: Physical*, vol. 167, no. 2, pp. 171–187, Jun. 2011, doi: 10.1016/j.sna.2011.02.038.

[6] S. Luo, J. Bimbo, R. Dahiya, and H. Liu, "Robotic tactile perception of object properties: A review," *Mechatronics*, vol. 48, pp. 54–67, 2017, doi: https://doi.org/10.1016/j.mechatronics.2017.11.002.

[7] G. Choi, M. G. A. Mohamed, and H. Kim, "Symmetric Signal Reconstruction and Frequency-Division Differential Driving for High Rate Touch Screen Sensing," *J. Display Technol.*, vol. 12, no. 11, pp. 1423–1432, Nov. 2016, doi: 10.1109/JDT.2016.2602244.

[8] A. H. Anwer *et al.*, "Recent Advances in Touch Sensors for Flexible Wearable Devices," *Sensors*, vol. 22, no. 12, p. 4460, Jun. 2022, doi: 10.3390/s22124460.

[9] P. Li, H. P. Anwar Ali, W. Cheng, J. Yang, and B. C. K. Tee, "Bioinspired Prosthetic Interfaces," *Adv Materials Technologies*, vol. 5, no. 3, p. 1900856, Mar. 2020, doi: 10.1002/admt.201900856.

[10] R. Lazzarini, R. Magni, and P. Dario, "A tactile array sensor layered in an artificial skin," in *Proceedings 1995 IEEE/RSJ International Conference on Intelligent Robots and Systems. Human Robot Interaction and Cooperative Robots*, Pittsburgh, PA, USA: IEEE Comput. Soc. Press, 1995, pp. 114–119. doi: 10.1109/IROS.1995.525871.

[11] M. Wang *et al.*, "Tactile Near-Sensor Analogue Computing for Ultrafast Responsive Artificial Skin," *Advanced Materials*, vol. 34, no. 34, p. 2201962, Aug. 2022, doi: 10.1002/adma.202201962.

[12] R. S. Dahiya, G. Metta, M. Valle, and G. Sandini, "Tactile Sensing—From Humans to Humanoids," *IEEE Trans. Robot.*, vol. 26, no. 1, pp. 1–20, Feb. 2010, doi: 10.1109/TRO.2009.2033627.

[13] W. W. Lee *et al.*, "A neuro-inspired artificial peripheral nervous system for scalable electronic skins," *Sci. Robot.*, vol. 4, no. 32, p. eaax2198, Jul. 2019, doi: 10.1126/scirobotics.aax2198.

[14] F. Bergner, P. Mittendorfer, E. Dean-Leon, and G. Cheng, "Event-based signaling for reducing required data rates and processing power in a large-scale artificial robotic skin," in *2015 IEEE/RSJ International Conference on Intelligent Robots and Systems (IROS)*, Hamburg, Germany: IEEE, Sep. 2015, pp. 2124–2129. doi: 10.1109/IROS.2015.7353660.

[15] C. Bartolozzi et al., "Event-driven encoding of off-the-shelf tactile sensors for compression and latency optimisation for robotic skin," in *2017 IEEE/RSJ International Conference on Intelligent Robots and Systems (IROS)*, Vancouver, BC: IEEE, Sep. 2017, pp. 166–173. doi: 10.1109/IROS.2017.8202153.

[16] W. L. Chan, M. L. Moravec, R. G. Baraniuk, and D. M. Mittleman, "Terahertz imaging with compressed sensing and phase retrieval," *Opt. Lett.*, vol. 33, no. 9, p. 974, May 2008, doi: 10.1364/OL.33.000974.

[17] P. K. Baheti and H. Garudadri, "An Ultra Low Power Pulse Oximeter Sensor Based on Compressed Sensing," in *2009 Sixth International Workshop on Wearable and Implantable Body Sensor Networks*, Berkeley, CA: IEEE, Jun. 2009, pp. 144–148. doi: 10.1109/BSN.2009.32.

[18] B. Hollis, S. Patterson, and J. Trinkle, "Adaptive basis selection for compressed sensing in robotic tactile skins," in *2017 IEEE Global Conference on Signal and Information Processing (GlobalSIP)*, Montreal, QC: IEEE, Nov. 2017, pp. 1285–1289. doi: 10.1109/GlobalSIP.2017.8309168.

[19] B. Hollis, S. Patterson, and J. Trinkle, "Compressed Learning for Tactile Object Recognition," *IEEE Robot. Autom. Lett.*, vol. 3, no. 3, pp. 1616–1623, Jul. 2018, doi: 10.1109/LRA.2018.2800791.

[20] B. Hollis, S. Patterson, and J. Trinkle, "Compressed sensing for tactile skins," in *2016 IEEE International Conference on Robotics and Automation (ICRA)*, Stockholm, Sweden: IEEE, May 2016, pp. 150–157. doi: 10.1109/ICRA.2016.7487128.

[21] B. Hollis, "Compressed Sensing for Scalable Robotic Tactile Skins," PhD Thesis, 2018. [Online]. Available: https://proxy1.library.jhu.edu/login?url=https://www.proquest.com/dissertations-theses/compressed-sensing-scalable-robotic-tactile-skins/docview/2085207014/se-2

[22] L. Shao, T. Lei, T.-C. Huang, Z. Bao, and K.-T. Cheng, "Robust Design of Large Area Flexible Electronics via Compressed Sensing," in *2020 57th ACM/IEEE Design Automation Conference (DAC)*, San Francisco, CA, USA: IEEE, Jul. 2020, pp. 1–6. doi: 10.1109/DAC18072.2020.9218570.

[23] L. E. Aygun et al., "Hybrid LAE-CMOS Force-Sensing System Employing TFT-Based Compressed Sensing for Scalability of Tactile Sensing Skins," *IEEE Trans. Biomed. Circuits Syst.*, vol. 13, no. 6, pp. 1264–1276, Dec. 2019, doi: 10.1109/TBCAS.2019.2948326.

[24] L. E. Aygun et al., "17.3 Hybrid System for Efficient LAE-CMOS Interfacing in Large-Scale Tactile-Sensing Skins via TFT-Based Compressed Sensing," in *2019 IEEE International Solid- State Circuits Conference - (ISSCC)*, San Francisco, CA, USA: IEEE, Feb. 2019, pp. 280–282. doi: 10.1109/ISSCC.2019.8662442.

[25] W. W. Lee, S. L. Kukreja, and N. V. Thakor, "Discrimination of Dynamic Tactile Contact by Temporally Precise Event Sensing in Spiking Neuromorphic Networks," *Front. Neurosci.*, vol. 11, Jan. 2017, doi: 10.3389/fnins.2017.00005.

[26] A. Slepyan, M. Zakariaie, T. Tran, and N. Thakor, "Wavelet Transforms Significantly Sparsify and Compress Tactile Interactions," *Sensors*, vol. 24, no. 13, p. 4243, Jun. 2024, doi: 10.3390/s24134243.

[27] S. Hsiao, "Central mechanisms of tactile shape perception," *Current Opinion in Neurobiology*, vol. 18, no. 4, pp. 418–424, Aug. 2008, doi: 10.1016/j.conb.2008.09.001.

[28] R. J. Watt, "Scanning from coarse to fine spatial scales in the human visual system after the onset of a stimulus," *J. Opt. Soc. Am. A*, vol. 4, no. 10, p. 2006, Oct. 1987, doi: 10.1364/JOSAA.4.002006.

[29] Y. Wu, M. Rosca, and T. Lillicrap, "Deep Compressed Sensing," in *Proceedings of the 36th International Conference on Machine Learning*, K. Chaudhuri and R. Salakhutdinov, Eds.,

in Proceedings of Machine Learning Research, vol. 97. PMLR, Jun. 2019, pp. 6850–6860. [Online]. Available: https://proceedings.mlr.press/v97/wu19d.html

[30] S. Wu *et al.*, "Learning a Compressed Sensing Measurement Matrix via Gradient Unrolling," in *Proceedings of the 36th International Conference on Machine Learning*, K. Chaudhuri and R. Salakhutdinov, Eds., in Proceedings of Machine Learning Research, vol. 97. PMLR, Jun. 2019, pp. 6828–6839. [Online]. Available: https://proceedings.mlr.press/v97/wu19b.html

[31] M. Aharon, M. Elad, and A. Bruckstein, "K-SVD: An algorithm for designing overcomplete dictionaries for sparse representation," *IEEE Transactions on Signal Processing*, vol. 54, no. 11, pp. 4311–4322, Nov. 2006, doi: 10.1109/TSP.2006.881199.

# Supplementary Material for:
# Single-Pixel Tactile Skin via Compressive Sampling

**Authors:** Ariel Slepyan[1]*†, Rudy Zhang[1]†, Laura Xing[2]†, Nitish Thakor[1,2,3]*

**Affiliations:**

[1]Department of Electrical and Computer Engineering, Johns Hopkins University, 3400 North Charles Street, Baltimore, MD 21218, USA.

[2]Department of Biomedical Engineering, Johns Hopkins School of Medicine, 720 Rutland Avenue, Baltimore, MD 21205, USA.

[3]Department of Neurology, Johns Hopkins University, 600 North Wolfe, Baltimore, MD 21205, USA.

*Corresponding author. Ariel Slepyan: aslepya1@jhu.edu. Nitish Thakor: nitish@jhu.edu

†These authors contributed equally to this work

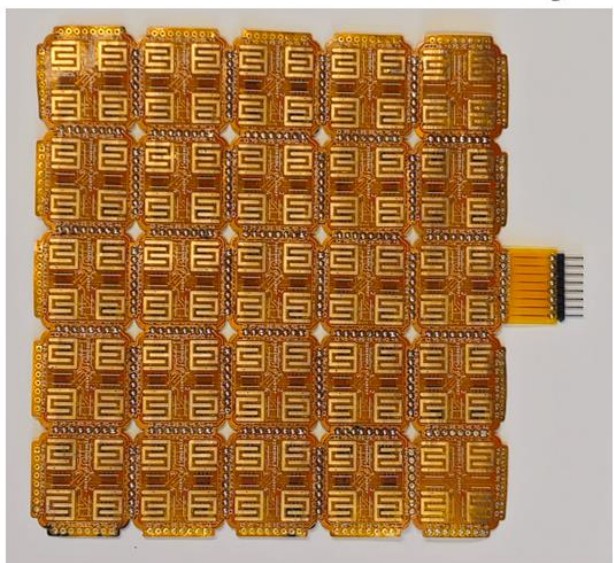
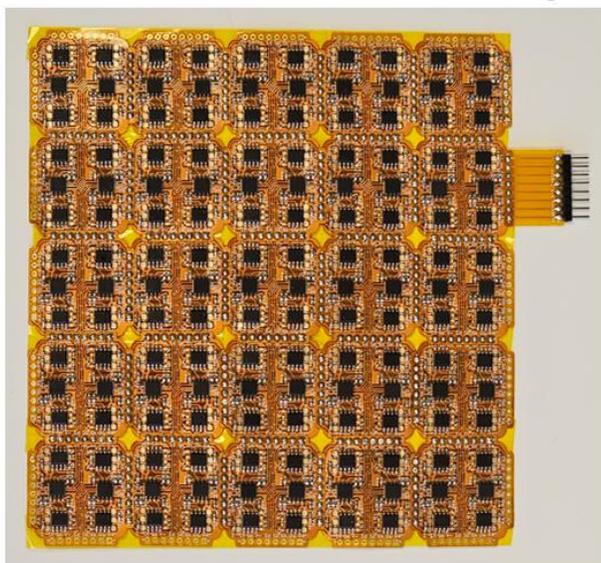
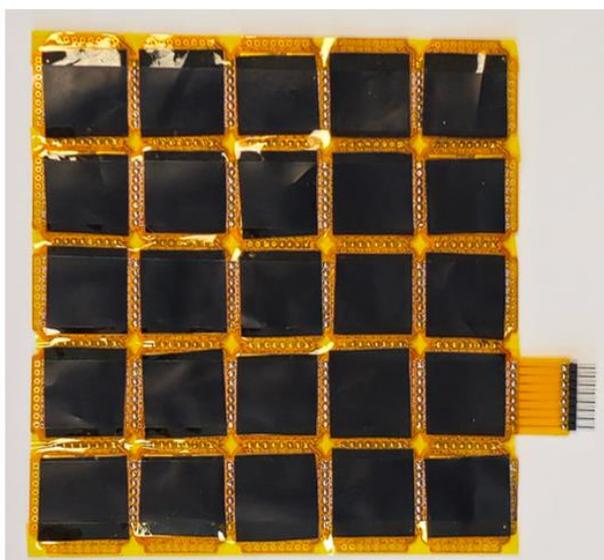

**Figure S1.** SPTS array assembly

**Figure S2.** SPTS cell schematic and board layout.

| Reference | Value |
|---|---|
| C1, C2, C3 | 0.1μF, 0805 |
| IC1 | ATtiny412 |
| IC2 | TLV9362 |
| J1 | 3* 1.5mm square pads, UPDI Programming |
| J2 | Conn_01x08_Pin, 2.54mm |
| R1 | Piezoresistive sensor (Variable by Pressure) |
| R2, R3 | 4,700 Ω, 0805 |
| R4, R5 | 1,000,000 Ω, 0805 |
| Piezoresistive Layer | Velostat |
| Protective Foam | Neoprene |

**Figure S3.** Component list for SPTS cell.

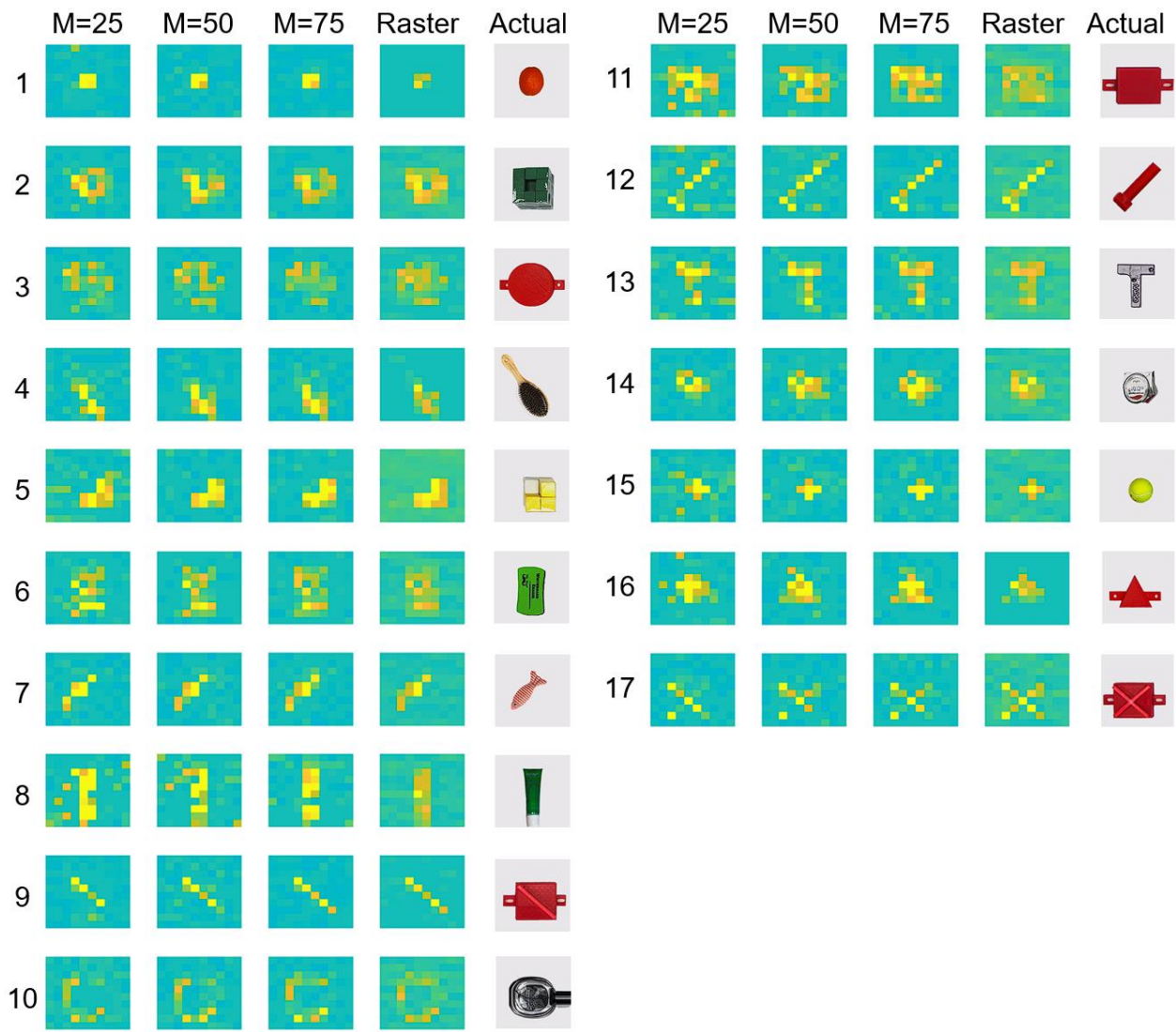

**Figure S4.** Example tactile reconstructions of SPTS at various measurement levels.

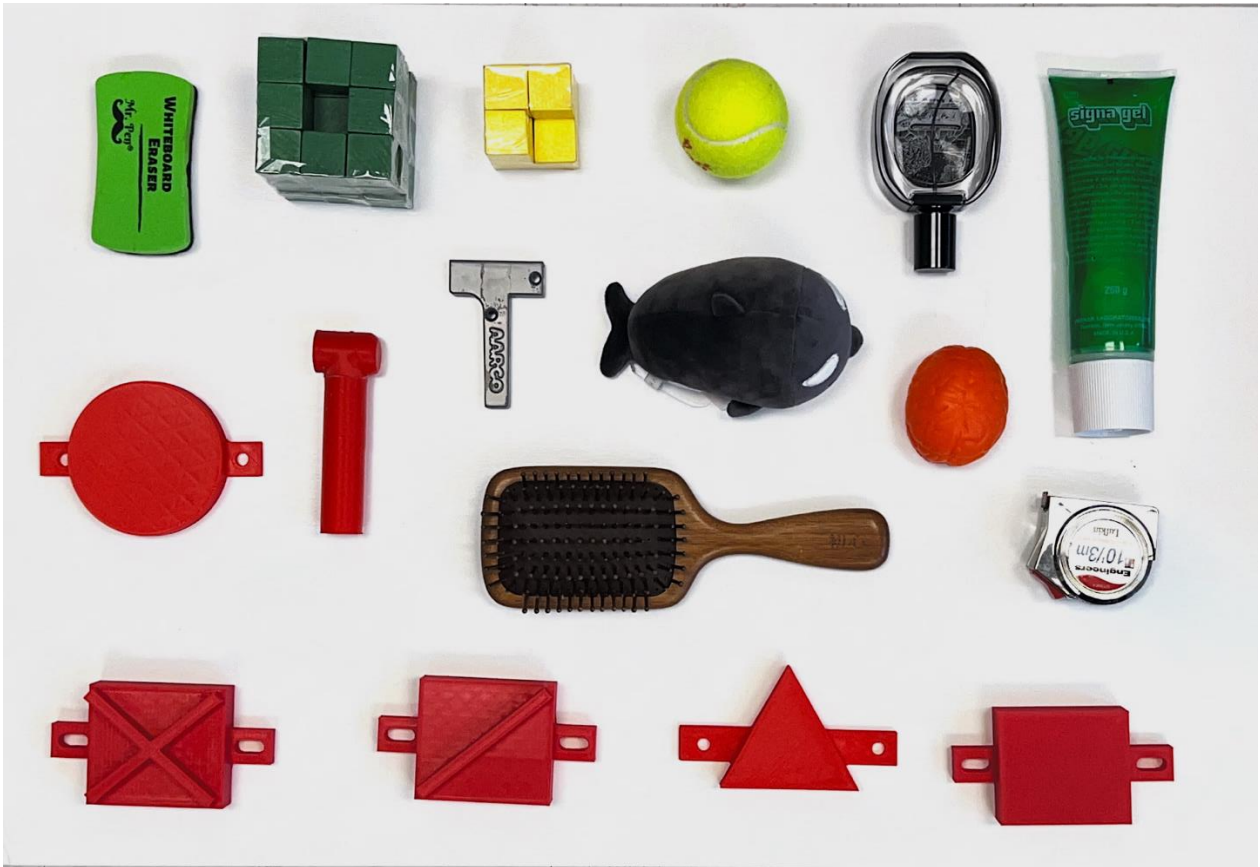

**Figure S5.** Picture of 17 objects used in object classification test.

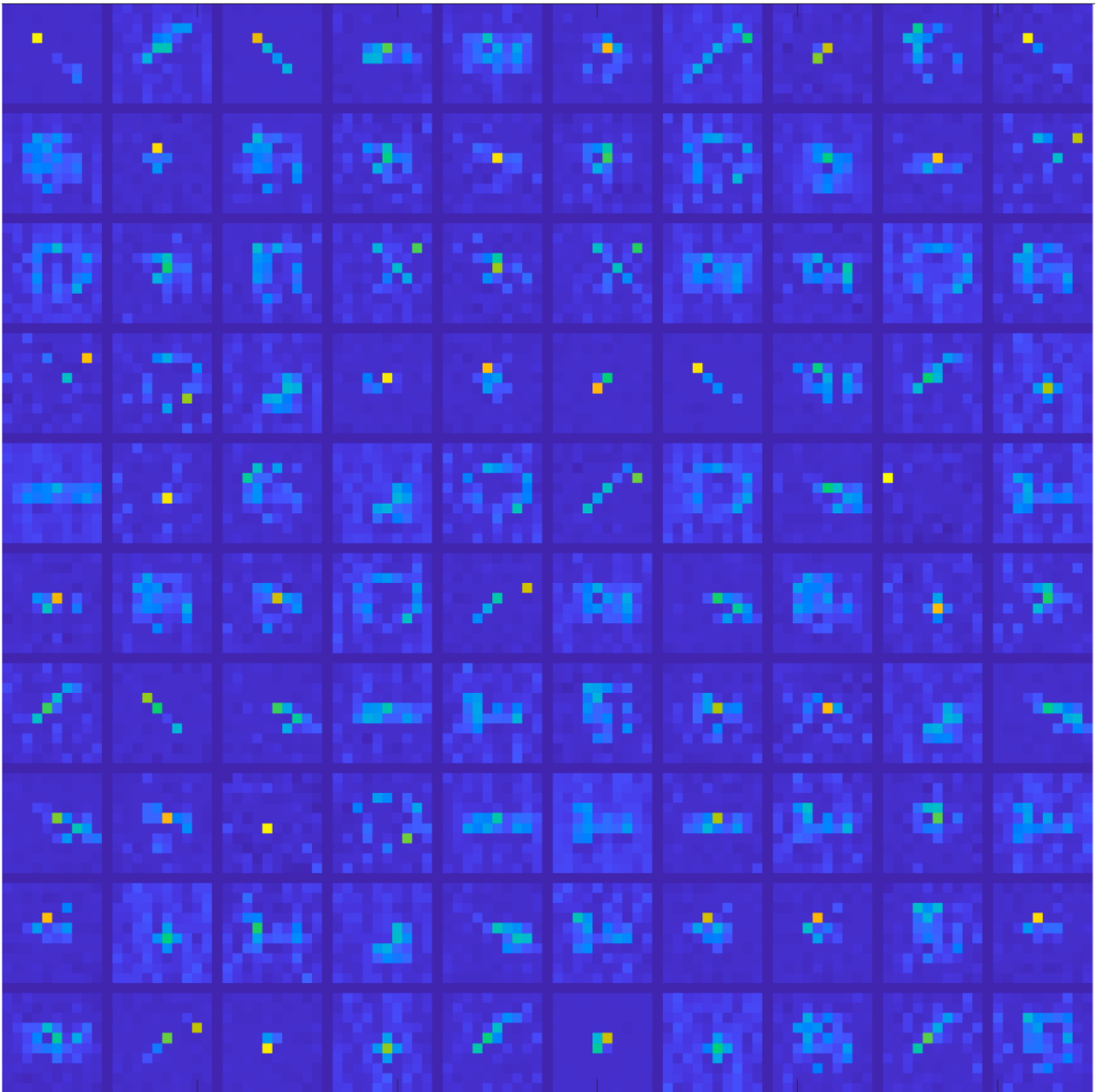

**Figure S6.** Learned tactile dictionary for sparse recovery of tactile signals.